\documentclass[twoside,11pt]{article}

\usepackage{jmlr2e}

\usepackage{amssymb}
\usepackage{amsmath}
\usepackage{todonotes}
\usepackage{graphicx}
\usepackage{float}
\usepackage{caption}
\usepackage{subcaption}
\ShortHeadings{Reinforcement Learning Based Online Decision Trees}{Garlapati, Raghunathan, Nagarajan and Ravindran}
\firstpageno{1}

\begin{document}

\title{A Reinforcement Learning Approach to Online Learning of Decision Trees}

\author{\noindent
       \name Abhinav Garlapati \thanks{The authors contributed equally} \email abhinavg@cse.iitm.ac.in %\\
       %\addr Indian Institute of Technology, Madras        \\
       \AND
       \name Aditi Raghunathan \footnotemark[1] \email aditirag@cse.iitm.ac.in %\\
       %\addr Indian Institute of Technology, Madras        \\
       \AND
       \name Vaishnavh Nagarajan \footnotemark[1] \email         vais@cse.iitm.ac.com %\\
       %\addr Indian Institute of Technology, Madras        \\
       \AND
       \name Balaraman Ravindran \email ravi@cse.iitm.ac.in \\
       \addr Deparment of Computer Science and Engineering\\
       Indian Institute of Technology, Madras\\
       }

\maketitle

\begin{abstract}%   <- trailing '%' for backward compatibility of .sty file
Online decision tree learning algorithms typically examine all features of a new data point to update model parameters. We propose a novel alternative, Reinforcement Learning-based Decision Trees (RLDT), that uses Reinforcement Learning (RL) to actively examine a minimal number of features of a data point to classify it with high accuracy.
Furthermore, RLDT optimizes a long term return, providing a better alternative to the traditional \textit{myopic} greedy approach to growing decision trees.
We demonstrate that this approach performs as well as batch learning algorithms and other online decision tree learning algorithms, while making significantly fewer queries about the features of the data points. We also show that RLDT can effectively handle concept drift.

\end{abstract}

\section{Introduction}
Decision trees sequentially test the values of various features of a data point to report a class label. The feature to be inspected at a stage is a decision that depends on the results of the tests conducted previously. It is natural to perceive this problem as a sequential decision-making task, which can be solved using Reinforcement Learning (RL). More importantly, when compared to traditional decision tree learning, we will see how the different constructs in RL offer better control over multiple aspects of the learning algorithm.
%Super line :)

%Changes (Aditi) 2nd May : Added RL like after sequential decision making
% Minimising the information not just while labeling but also while learning, tried to make that explicit
%for - an in MDP
%A change in the line about different costs

In this work, we present {RLDT}, an RL-based online decision tree algorithm for classification. Our objective is to design a learner that can achieve high accuracy while simultaneously minimizing the amount of information about data points used for both predicting \textit{and} learning. %required to label a data point.%
We show that learning such a minimal tree is equivalent to solving for the optimal policy of an unknown Markov Decision Process (MDP).

%Any decision tree incurs a cost of querying a feature and obtains a reward at the end of all its tests when it decides to report a label.

A challenge in searching over the space of decision trees is the combinatorial complexity of the problem. Traditional algorithms address this by taking a greedy decision while splitting any node (e.g., splitting along feature with highest information gain) to grow the tree. As shown by \citet{greedy}, this can lead to sub-optimal solutions. The decision making problem could be solved better by maximizing a \textit{long-term return} rather than a short term gain - which is a typical goal in RL.

%\citet{greedy} show that this is not necessarily optimal.
%This may not necessarily be optimal \citep{greedy}. What's wrong with current sentence? Length? I kulted line from some other paper :p

Sometimes decision trees can have very few training points in their leaf nodes. This overfitting is usually avoided by pruning the tree. RLDT, however, does not have an explicit pruning stage. The RL formulation provides a neat way of striking a balance between pruning for generalization and growing deeper trees for higher accuracy.

%By penalising the queries, we make the learning algorithm make as few queries to features as possible, without compromising on the accuracy.
%The requirement of having to ask as few questions as possible (about the data point) can also be motivated from the practical standpoint that data is usually \textit{expensive}.
%%%%%%%%%%%%% I am not sure this line is necessary - we need to save space
%Often, data come with multiple features all of which may not be relevant to the task at hand.

We are interested in a learner that asks as few questions as possible about the features of the data point since data can be expensive. Furthermore, different features could have different costs associated with their extraction. We would want to selectively and efficiently examine values subject to these parameters. We will see that the reward formulation of the MDP can be naturally tuned to learn in this setup.

% As explained ahead, it is straightforward to incorporate this given the the reward framework of RL
%This can be easily incorporated as rewards in the RL framework.

%We are interested in the process of learning and converging to the the optimal decision tree for the given data distributions and feature costs, as we get more points, incurring maximum reward in the process.
Our interest in studying an RL based approach is also motivated by the fact that online classification models face the challenge of concept drift, where the optimal tree changes with time. Online RL algorithms efficiently handle such situations with non-stationary optimal solutions.

Finally, we note that the above formulation is rich in that any of the vast variety of RL algorithms can be used to learn the optimal policy which forms our decision tree. Thus RLDT promises different techniques for online decision tree learning within a single framework.
%Another super sentence :)

The rest of the paper is organized as follows. In Section \ref{sec:earlier}, we discuss related work in online decision tree learning. Section \ref{sec:problem} formally defines the problem while Section \ref{sec:mdp} and \ref{sec:solve} present the MDP formulation in RLDT. We discuss experiments in Section \ref{sec:exp} after which we discuss possible directions for future work.

\section{Earlier Work}
\label{sec:earlier}
%The interpretability of decision trees (also called as classification trees) \citet{DecisionTrees} make them very popular classification models.

Typically, online decision tree learning algorithms such as {VFDT}  \citep{VFDT} and {ID4}  \citep{ID4} try to incrementally learn the same tree that batch training algorithms like C4.5 learn. They examine all the features of the input data point to update their models. However RLDT uses a minimal subset of these features to both classify a new data point and to update its parameters.

Our work is closely related to that of \citet{ofs} in which the authors study an online linear classification algorithm that examines a fixed number of features to classify an input point. As the authors note, the constraints of this algorithm are harder than that in sparse online learning \citep{sol1, sol2} that has no upper bound on the number of features examined. RLDT, in fact, imposes a harder combination of the constraints as the learner is encouraged to examine the fewest number of features, subject to an upper bound.
%peace
%:)

We must note how RLDT differs from earlier studies of RL-based decision tree learning. \citet{pomdp} model the process of learning decision trees as a \textit{known} partially observable MDP  where the states of the underlying MDP are the training sample points. The learner decides to query a feature/report a label based on its belief over these states; the belief encodes the knowledge about the point that has been acquired so far. \citet{mdp} approach the problem by designing an MDP with states corresponding to every possible partition of the training data. The action defined at a state is any kind of test (a feature query) that can be conducted on {one of the subsets} in that partition of the sample points.

%of the above approaches? or with (Aditi)?
A serious drawback of the above approaches is that the solutions are very rigid with respect to the training data. The algorithms fail to work in an incremental setup because the state spaces change completely by the addition of a new point. RLDT on the other hand is different in that it uses a state space representation that allows learning from new data incrementally. While the algorithm presented by \citet{selforg} uses a similar representation, it is designed for learning from an offline batch of data points and an extension to online learning is not obvious. Furthermore, \citet{selforg} propose a reward parametrization that is not concerned with minimizing the number of queries.

\section{Online Learning for Classification with Query Costs}
\label{sec:problem}
% Input space is categorical for now? - Yes
We now formally define the problem of learning to classify online by querying on features selectively. We assume that the input space is $\mathcal{X} = \mathbb{Z}^d$ and the output space is a set of labels $\mathcal{Y}=\left\lbrace 1, 2, \hdots K\right\rbrace$. We discuss an extension to continuous features in the next section. At every timestep $t = 1, 2, \hdots$, the environment draws an input $\vec{x}_t = (x_1, x_2, \hdots x_d)\in \mathcal{X}$ and output $y_t \in \mathcal{Y}$ according to an underlying distribution $\mathcal{D}$. The learner is initially not informed of any of the attributes of $\vec{x}_t$. Instead, the learner interacts with the environment and \textit{actively} informs itself about the point in two ways.
%in two ways -- it informs itself of the features in only one way no (Aditi)?
First, it can either \textit{\textbf{query}} the environment about the value of a feature with index $j$ -- which is denoted as executing action $\mathcal{F}_j$. Secondly, the learner can \textit{\textbf{report}} to the environment an answer $\hat{y}_t \in \mathcal{Y}$ which is denoted as action $R_{\hat{y}_t}$. The environment then reveals the true label $y_t$, provides a reward depending on whether $\hat{y}_t = y_t$ or not, and considers the episode (timestep) terminated.
%Should we mention timestep, cause it suddenly changed from timestep to episode (Aditi)
The objective of the learner involves a) achieving a good performance in classification and b) making few queries.
%Aditi in/of classification?

\section{MDP Formulation}
\label{sec:mdp}
We formulate the above problem as a Markov Decision Process $M$ such that each episode of the MDP processes a new input point. The state space $\mathcal{S}$ of the MDP  consists of all possible states that encode partial information about the data point. This partial information consists of the values of the features that are considered to be known at that state. Thus, every state $s \in \mathcal{S}$ corresponds to a set of known features $f(s)$ which have a configuration unique to $s$. Clearly, for the initial state $s_0$ of the MDP, $f(s_0)= \phi$.

The action space $\mathcal{A}$ consists of the feature query actions $\mathcal{F}_i$ for $i = 1, 2, \hdots d$ and the report actions $\mathcal{R}_i$ for $i = 1,2, \hdots K$. However, not all query actions are allowed on all states. Consider feature $j \in f(s)$. $\mathcal{F}_j$ is disallowed at $s$. Furthermore, on taking action $\mathcal{F}_i$ at this state ($i\notin f(s)$) a transition is made to another state $s'$ that includes all partial information present in $s$ and also the information about the value of feature $i$. Thus $f(s') = f(s) \cup \{i\}$. On the other hand, taking action $\mathcal{R}_k$ leads us to a state where the label of the point is known regardless of which report action was taken.
While here we consider only queries that result in as many splits as the number of values the feature can take, note that one could also consider comparison queries that result in binary splits.

The rewards on the query actions $\mathcal{F}_i$ are negative and are proportional to the cost  $-C_{i}$ of querying feature $i$ of the point $\vec{x}_t$. The reward on the report action $\mathcal{R}_i$ is however dependent on the actual label $y_t$ and the reported label $i$. If $y_t = i$, the agent assumes that the environment reinforces it with a positive reward, and a negative reward otherwise.

In order to discourage the agent from asking many queries we have to set non-zero query costs ($C_{i}>0$) \textit{or/and} lower values of $\gamma$, the discount factor.
 While $\gamma$ discourages querying independent of the feature being queried, query costs provide us more flexibility to encode the actual cost associated with accessing a feature.
%Added stuff here (Aditi)

\section{Solving the MDP}
\label{sec:solve}

%We employ Q-learning \citet{Qlearning} to solve the MDP $M$ formulated as above
%\textbf{Q-learning}  is a popular model-free method to solve MDPs.
We employ Q-learning to solve the MDP $M$ formulated as above. We chose Q-learning (off-policy method) over SARSA (on-policy method) because we perform multiple path updates (described in Section \ref{sec:backup}).
The details of the process is described below and experimental results follow.

\subsection{The reward for classifications and misclassification}
When the true label is revealed after a report action, the value estimate of the report action for the true label is updated with a target of $R_+$ while the report action for each of the other labels is updated with a target of $R_-$. At some state $s$, for a class $k$, assume that $\mathcal{D}$ is such that any point that is consistent with the partial information at that state belongs to class $k$ with probability $p$. Thus, we know that  $Q(s,\mathcal{R}_k)$ converges to the expected return of $$ pR_+ + (1-p)R_- .$$

\vspace{-1pt}

Consider a query action $\mathcal{F}_i$ at $s$ for which state $t$ is one of the possible subsequent states. Given state $t$, it is likely that there exists a class $k$ such that its density $p_k$ in state $s$, is less than its density  $p_k'$ in $t$, since we have more information about the point in $t$. Thus, $p_k' > p_k$. Taking action  $\mathcal{F}_i$ and reporting $k$ in state $t$ results in an expected return of $$-C_{i} + \gamma (p_k'R_+  + (1-p_k') R_-).$$ \vspace{-1pt} The amount by which this expected return differs from the expected reward on directly taking $\mathcal{R}_k$ at state $s$ would be $$
-C_{i} + (\gamma  p_k' - p_k)R_+ + (- 1+\gamma - \gamma p_k' + p_k) R_- .
$$\vspace{-1pt}

The above quantity can be made both positive and negative depending on the parameters $\gamma, C_{i}, R_+, R_-$. This is precisely how the tradeoff between accuracy and number of queries is addressed in RLDT. Furthermore, carefully setting the values $R_+$ and $R_-$ to be a function of the label being correctly/wrongly reported, we can enforce cost-sensitive learning, and handle {class imbalance}. Class imbalance can also be addressed by setting the value of $\alpha$ to be lower for the majority class thereby under-sampling the class.
%CHanged aditi

\subsection{Speeding up Convergence}
\label{sec:backup}
We say that RLDT converges when it has seen sufficiently many points beyond which the algorithm's performance does not change in expectation as it has converged to the optimal policy/decision tree. In this section, we discuss two techniques that will help in faster convergence.
%In practice, we would require the algorithm to converge to an optimal solution and subsequently adapt to concept drift within a small number of time steps. The following methods prove to be useful in speeding up convergence.

\begin{figure}[H]
\centering
\includegraphics[scale=0.7]{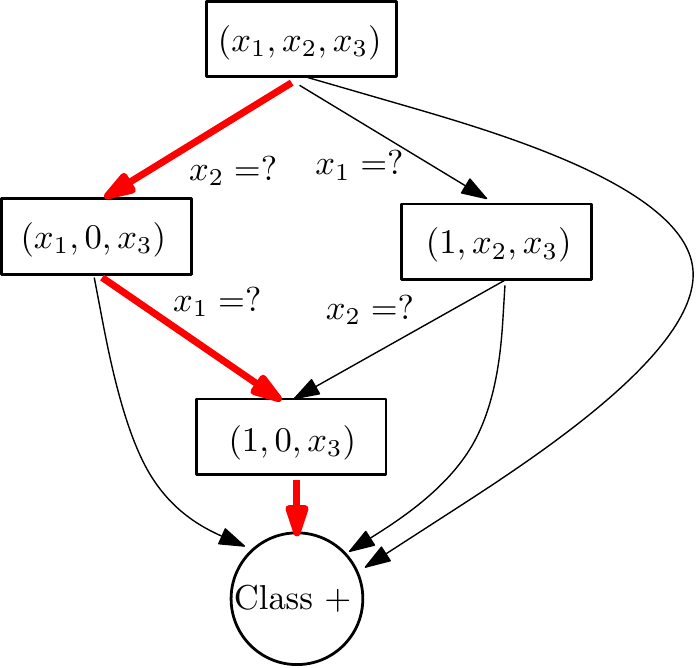}
\caption{Paths along which updates must be made}
\label{fig:backup}
\end{figure}

\subparagraph{Multiple Path Updates}
The updates after every episode correspond to a specific order in which the queries were asked. This order corresponds to a single `path' down the MDP. However, it is crucial to note that we could have made the same set of queries in any other order and effected these updates in other paths. We could have also taken shorter paths that made only a subset of these queries before reporting the label. As an example, in Figure \ref{fig:backup}, while the red edges correspond to the path that was taken, value updates can be performed on all the  action-values that have been shown. These updates must be done exactly once per state-action pair to avoid bias. We will see that this makes convergence significantly faster. We note that to avoid enumerating the exponential number of paths, we could sample a constant number of random paths from the space of paths to perform the updates. This would still be speed up convergence when compared to the naive single path update.

%Hence, on the input point, we have queried the value of $x_2$ first, and then the value of $x_1$ and finally reported the positive label. The only other order of querying that is possible for this case is querying about $x_1$ followed by $x_2$. What we propose here is that, the state-action pairs on this path too must be undergo one update each.

This technique however results in more frequent updates in states where fewer features about the point are considered to be known i.e.,  when $|f(s)|$ is small. Thus the value of the best report actions at these states converge faster than possibly better query actions. Since $\epsilon$-greedy exploration may not sufficiently explore these query actions, we use optimistic initial values for all the query actions. %When the initial values are set higher than the maximum value that a state can take, they serve as a simple way to encourage the query actions to explore more.
%Aditi - should we explain a bit more about optimal initial values?

%\begin{remark}
%It must be noted that this does not skew the learnt distribution in any sense because every state-action pair is allowed to update only once and this update reflects the fact that we saw the input point once within the part of the input space that corresponds to this state action pair.
%\end{remark}

%A further point here is that it is not just the paths containing permutations of all the queries that we must be considering for the $Q-$value updates. We could repeat the same with paths that correspond to making a subset of these queries. In the above example, this could be just reporting $+$ on the start state; or reporting the label after querying on $x_1$ or $x_2$. We show in our results that this technique helps in significantly increasing the convergence rate.

%\subsection{Random Backup Reuse}

\subparagraph{Truncated Exploration}

%\subsection{Non Exploratory Report Actions}
%While learning an optimal policy in an unknown MDP, the agent takes exploratory actions to acquire more information about the environment. In this context, we note that
Assume that we skip taking an exploratory report action $\mathcal{R}_i$ at a state $s$ and choose to make a query instead. At the end of the episode we would always take a report action and know the true label of the input point. By knowing the true label, we would also know the outcome of action $\mathcal{R}_i$ that we skipped at state $s$. Thus, the report actions at $s$ can be updated without exploring any of them. Hence, we restrict exploratory steps to only the query actions.%., and this truncated exploration results in faster convergence and better overall performance.

\section{Experiment}
%We evaluate our approach on multiple datasets from the UCI repository. We chose datasets having categorical features and seemingly sufficient number of data points for our algorithm to converge to a solution.
%I don't think we need to go into so much detail about each dataset.

\subsection{Mushroom Dataset}
\label{sec:exp}

%We evaluate our approach on multiple datasets from the UCI repository. We chose datasets having categorical features and seemingly sufficient number of data points for our algorithm to converge to a solution.
%I don't think we need to go into so much detail about each dataset.

%Changed compare to evaluate Aditi
Since it is sufficient to either have a low value of $\gamma$ or a strictly positive query cost, we choose to set $\gamma = 0.8$ and all query costs zero. The report rewards are set as $R_+ = 5$ and $R_- = -5$. Apart from an exploration rate $\epsilon = 0.01$, we initialize all query action values to an optimistic value of $8$ (an arbitrarily chosen value that is slightly higher than the maximum reward of 5). We allow the learner to make at most three queries in order to truncate the state space of the MDP. We show that under this restriction RLDT can perform better than other algorithms.

%We allow the agent to query at most $3$ (out of $8$) to highlight its classifying ability even with explicit restriction in the number of queries during learning. Restricting the number of queries also helps to reduce the state space to allow faster convergence.
%\subparagraph{Performance on Mushroom Data Set}

\begin{figure}[h!]
\centering
        \begin{subfigure}[b]{0.3\textwidth}
                \includegraphics[width=\textwidth]{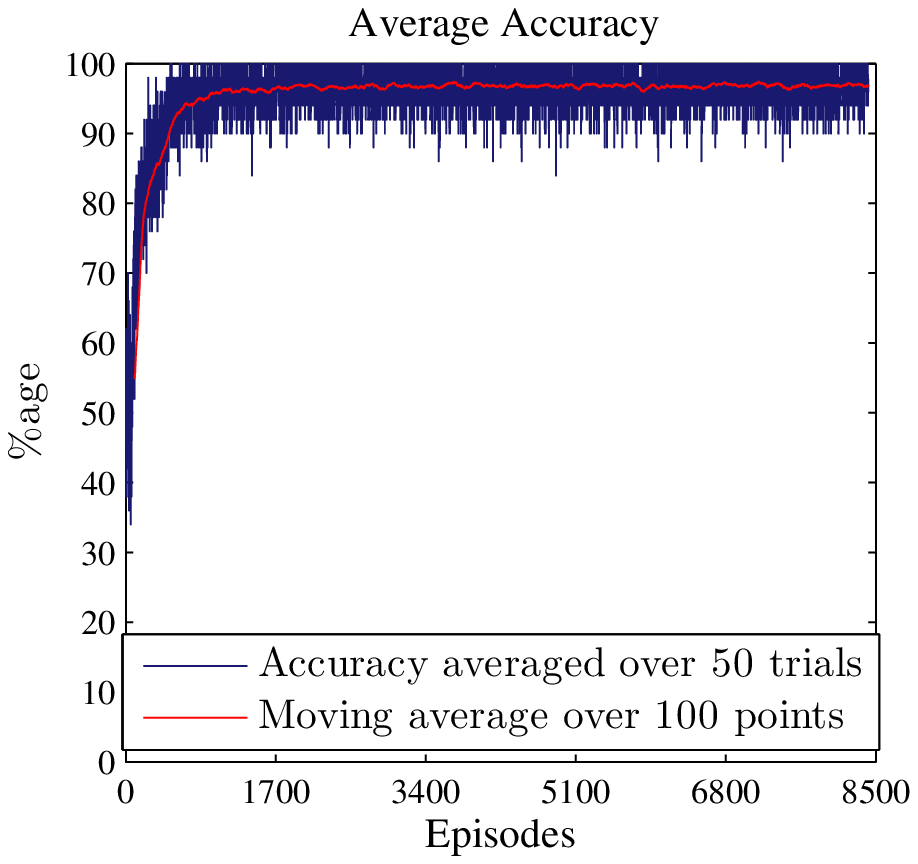}
                \caption{Average Accuracy}
                \label{fig:mush_acc}
        \end{subfigure}%
        ~ %add desired spacing between images, e. g. ~, \quad, \qquad, \hfill etc.
          %(or a blank line to force the subfigure onto a new line)
        \begin{subfigure}[b]{0.3\textwidth}
                \includegraphics[width=\textwidth]{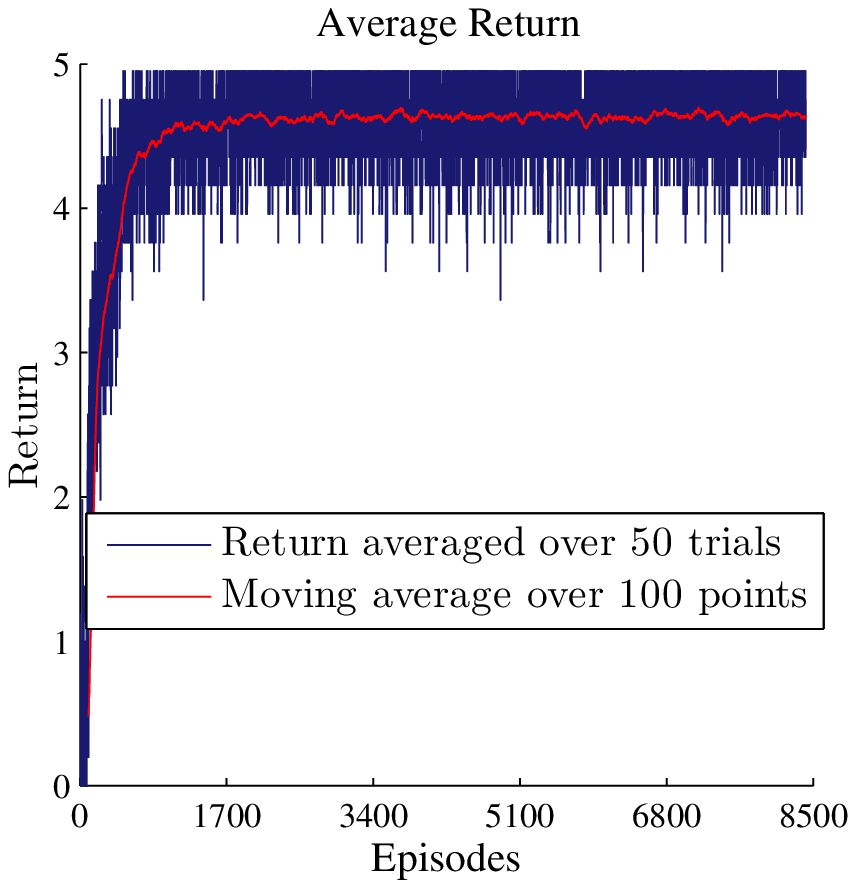}
                \caption{Average Return}
                \label{fig:mush_ret}
        \end{subfigure}
        ~ %add desired spacing between images, e. g. ~, \quad, \qquad, \hfill etc.
          %(or a blank line to force the subfigure onto a new line)
        \begin{subfigure}[b]{0.3\textwidth}
                \includegraphics[width=\textwidth]{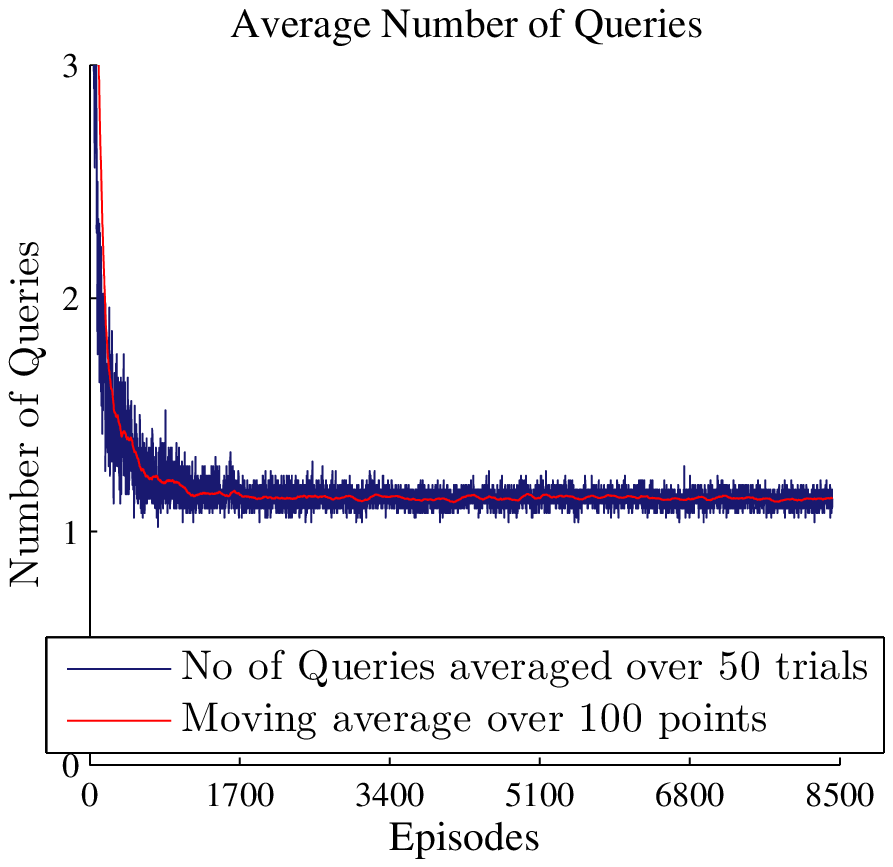}
                \caption{Average no. of Queries}
                \label{fig:mush_nQ}
        \end{subfigure}
        \caption{Performance on Mushroom Dataset with Multiple Path Updates}\label{fig:mushroom_backup}
\end{figure}

Figure \ref{fig:mushroom_backup} shows the graphs corresponding to the performance of RLDT averaged across 50 runs on the Mushroom Data Set \footnote{Available at \url{https://archive.ics.uci.edu/ml/datasets/Mushroom}}. Note that it is the return, which is a function of both accuracy and the number of queries, that the agent tries to optimize. We see that the learner attains a high accuracy after examining at most $3$ out of $22$ features of the first 1000 points. Beyond this, the number of queries the agent makes drops to $1.1$ queries per point in expectation  while maintaining an accuracy of $97.74\% \pm \textbf{3.01}$. This demonstrates that the discount factor discourages querying without affecting the accuracy undesirably. Note that all values are reported with their $95\%$ confidence intervals.

\begin{figure}[h!]
\centering
        \begin{subfigure}[b]{0.3\textwidth}
                \includegraphics[width=\textwidth]{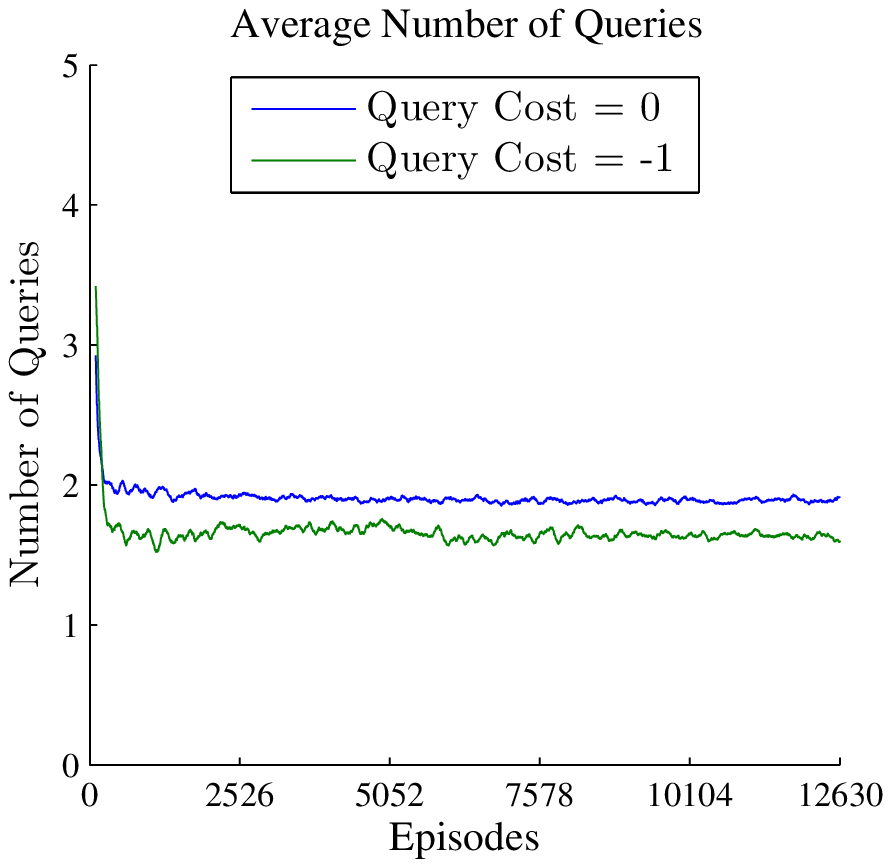}
                \caption{Varying Query Costs on Nursery Data Set}
                \label{fig:QueryCost}
        \end{subfigure}%
        ~ %add desired spacing between images, e. g. ~, \quad, \qquad, \hfill etc.
          %(or a blank line to force the subfigure onto a new line)
        \begin{subfigure}[b]{0.3\textwidth}
                \includegraphics[width=\textwidth]{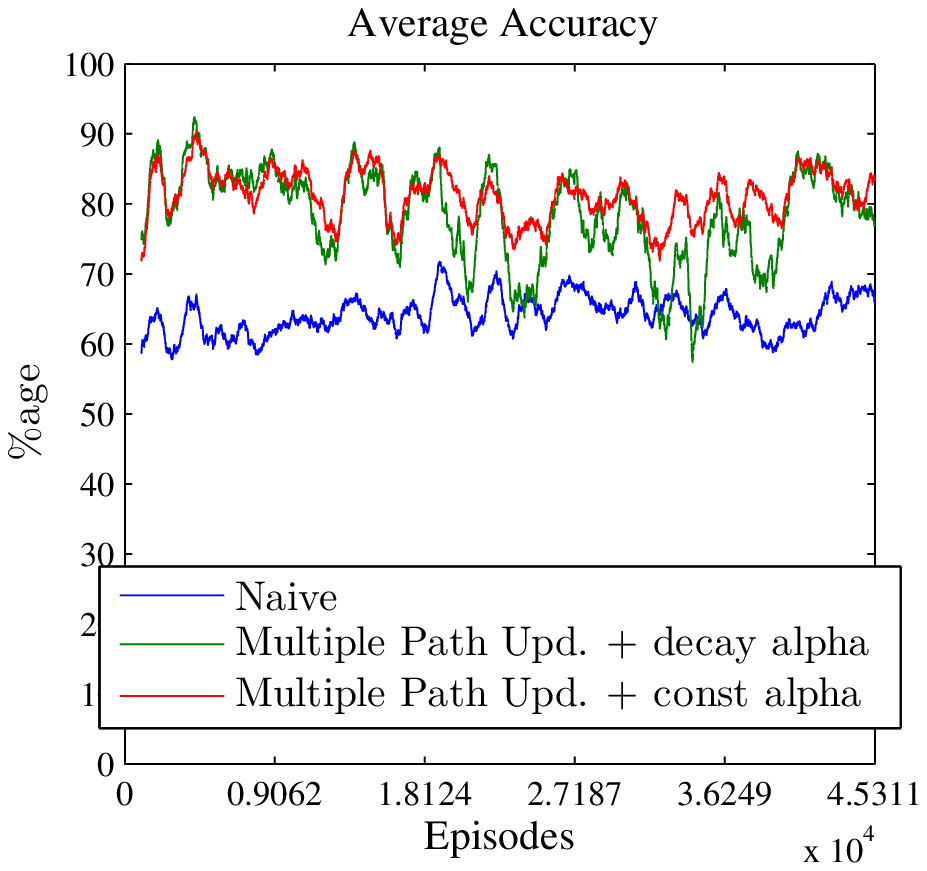}
                \caption{Average Accuracy on Electricity Data Set}
                \label{fig:elec_acc}
        \end{subfigure}

        \caption{}\label{fig:elec_plots}
\end{figure}

On the Nursery Data Set \footnote{Available at \url{https://archive.ics.uci.edu/ml/datasets/Nursery}} we only present the moving averages to observe the effect of the techniques discussed in Section \ref{sec:backup}. We see that multiple path updates speed up convergence significantly. Also, the \textit{truncated exploration} technique results in many queries being asked on the first few points as we would expect. However, a significant difference in the convergence rate is not noticeable in this case. On this data set, we also show that increasing query cost reduces the number of queries made during learning (Figure  \ref{fig:QueryCost}), although reducing accuracy from $92.53 \pm \textbf{1.22}\% $ to $88.5 \pm \textbf{0.88} \% $. We also observe that when the number of allowed queries is limited, RLDT can perform better than even a batch learning algorithm like C4.5 that learns from all features. On performing 5-fold cross validation, RLDT achieves an accuracy of $92.53\pm \textbf{1.22}$\% with $2$ feature queries per point in the testing phase. When C4.5 trees are restricted to a depth of $2$ using Reduced Error Pruning, the accuracy on the same five folds turns out to be $85.33 \pm \textbf{0.87}$\%.
% \ref{fig:Imbalance} shows how RLDT can handle class imbalance with weighted rewards. We modified the Nursery dataset to have $4000$ points of one class and $1000$ each of the other. With rewards set to $5, 10, 10$ respectively, for each class, we were able to increase the average F-score.

%This demonstrates that the agent converges to a policy that makes minimal number of feature queries, because $\gamma$, the discount factor discourages the agent from making more queries than necessary, without affecting the accuracy.\\

%\subparagraph{Performance on Nursery Data Set}

%Figure \ref{fig:nursery_plots} shows the performance of the different versions of RLDT on the Nursery Data Set \footnote{Available at \url{https://archive.ics.uci.edu/ml/datasets/Nursery}}.

 % shows the number of queries made for different query costs. As is evident, for a higher cost, fewer queries are made although compromising the accuracy ($88.5\%$ with query cost $-1$).

%However, with higher query costs, the accuracy is lowered (to $88.5\%$ with query cost $-1$) demonstrating the trade off between accuracy and number of queries discussed earlier.
 %It's also seen that truncated exploration leads to slightly higher returns.

We use the Electric Power Consumption Data Set \citep{elec} \footnote{Continuous attributes were discretized.} to examine the performance of RLDT under concept drift. As seen in Figure \ref{fig:elec_acc}, a sufficient learning rate $\alpha$ is required to maintain a tree that adapts to a changing distribution. RLDT performs better than most popular online decision tree learning algorithms (Table \ref{tab:elec}), while making significantly fewer queries.

\begin{table}[h!]
\small
\centering
\begin{tabular}{|c||c|c|}
\hline
 Method & Final Accuracy(\%) & Mean Accuracy(\%) \\
 \hline
 StARMiner Tree(ST) & 81.3 & 78.93\\
 Automatic StARMiner Tree(AST) & 80.1 & 79.15\\
 VFDT & 75.8 & 74.64\\
 VFDTcNB & 77.3 & 77.16\\
 RLDT & \textbf{83.26} & \textbf{81.15}\\
 \hline
\end{tabular}
\caption{Performance of standard algorithms on Electric Power Consumption Data Set}
\label{tab:elec}
\end{table}

%This datasets provides 8124 samples corresponding to different species of mushrooms, which are to be classified as edible and poisonous. Each point has 22 nominal features, of varying number of categories.

\begin{figure}[h!]
\centering
        \begin{subfigure}[b]{0.3\textwidth}
                \includegraphics[width=\textwidth]{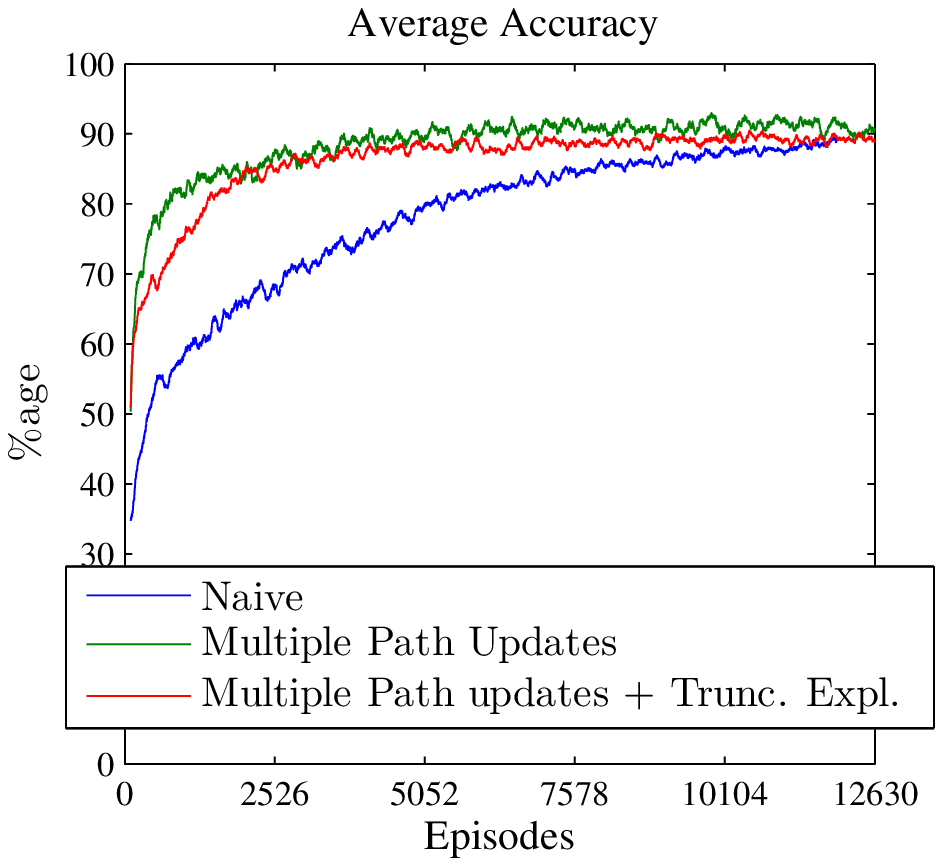}
                \caption{Average Accuracy}
                \label{fig:nursery_acc}
        \end{subfigure}%
        ~ %add desired spacing between images, e. g. ~, \quad, \qquad, \hfill etc.
          %(or a blank line to force the subfigure onto a new line)
        \begin{subfigure}[b]{0.3\textwidth}
                \includegraphics[width=\textwidth]{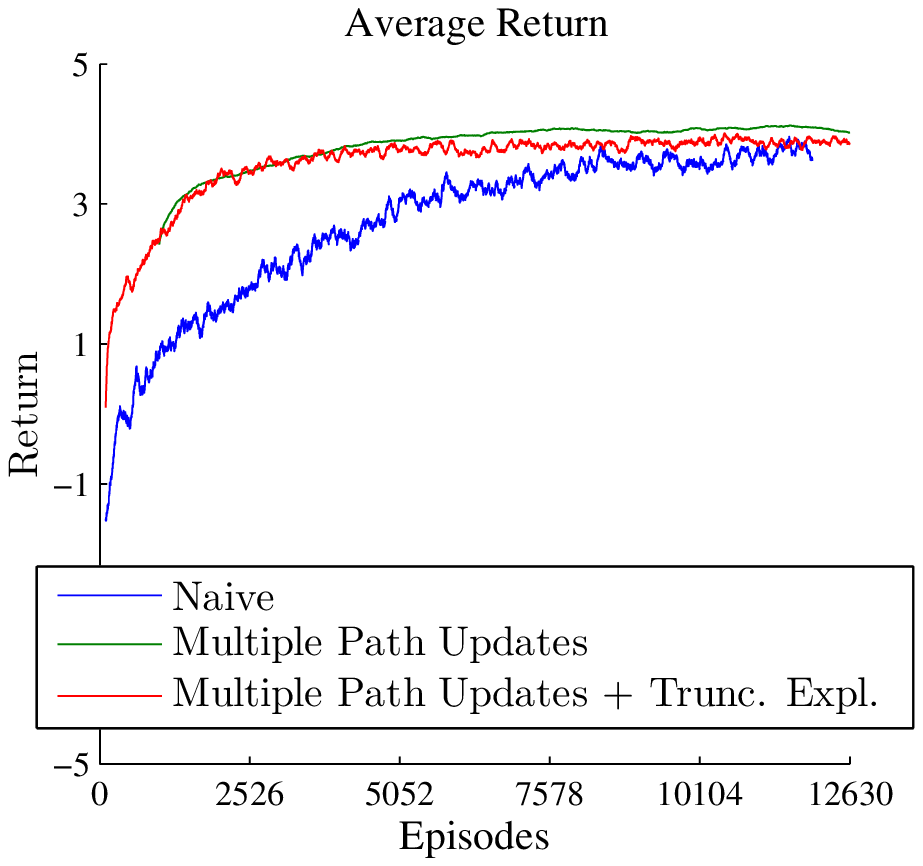}
                \caption{Average Return}
                \label{fig:nursery_ret}
        \end{subfigure}
        ~ %add desired spacing between images, e. g. ~, \quad, \qquad, \hfill etc.
          %(or a blank line to force the subfigure onto a new line)
        \begin{subfigure}[b]{0.3\textwidth}
                \includegraphics[width=\textwidth]{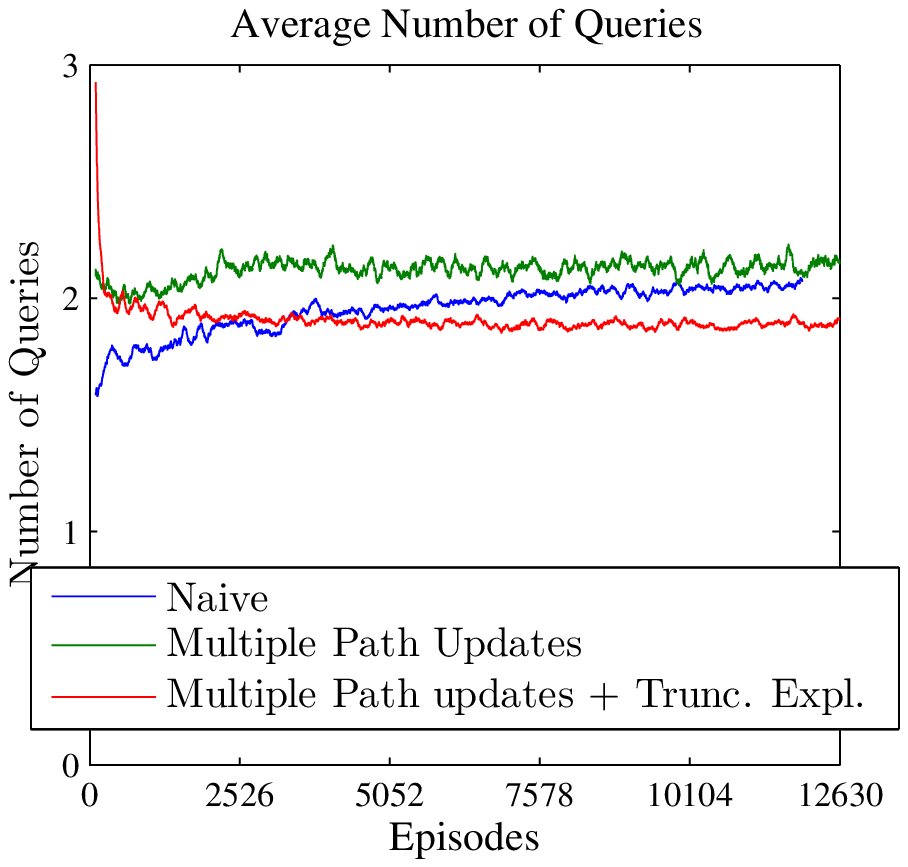}
                \caption{Average No. of Queries}
                \label{fig:nursery_nQ}
        \end{subfigure}
        \caption{Performance on Nursery Data Set}\label{fig:nursery_plots}
\end{figure}

\section{Future Extensions}
\label{sec:future}

In this section, we discuss possible ideas for developing RLDT to work with high-dimensional and continuous feature spaces.
%First, we would like to extend RLDT to handle large state spaces and action spaces. Next, the algorithm should also be able to handle continuous action spaces that correspond to continuous real-valued attributes.

\subparagraph{Dealing with large state and action spaces}
The MDP corresponding to high dimensional data will have large state and action spaces. Limiting the number of queries is one solution to truncating a large state space. We could also dynamically prune the query actions at a state if we observe that the actions lead to `equivalent' states where the best action is the same report action. The idea is to explicitly avoid exploration when not required.

On the other hand, we could use function approximation by encoding the states in terms of the features that are known/unknown at that state. However, efficient generalization can be done only with appropriate assumptions on the data distribution. For two states $s$ and $s'$, if neither of $f(s)$ and $f(s')$ is a subset of the other, the distribution in the part of the input space corresponding to  $s$ and $s'$ may differ too significantly that we may not be able to draw experience from $s$ to $s'$ or vice versa. Nevertheless, we  could still generalize if one of $f(s)$ or $f(s')$ is a subset of the other. That is, we can always draw experience from `sub-states' where more feature values are known, to `super-states' where less values are known. Two challenges arise here. First, we will have to determine how much weight experiences from different sub-states are given. Second, we would want to design a function approximation architecture that can (roughly) capture the overlapping experience between sub-states and super-states.

It is also important for the method to scale to large \textit{action spaces} which might otherwise hinder techniques like Q-learning. Policy gradient versions of RLDT with appropriate policy parametrization to handle this would be an interesting direction for future work. %would be a suitable way to solve this problem, given an appropriate parametrization.
%The large action space could hinder the performance of Q-learning.
%Added stuff here : Aditi%

\subparagraph{Continuous attribute values}
One way to handle continuous attribute values would be to discretize the feature space into finitely many bins. Alternatively, we could directly employ RL techniques that work with continuous action spaces. The value of splitting a continuous feature as a function of the split point is likely to be a continuous multimodal distribution which can be estimated online using a Gaussian Mixture Model \citep{gmm}.

Another alternative would be to maintain a finite number of split points per feature, each of which is dynamically updated as described by \citet{dynamic}. As the allowed set of actions changes gradually, the value estimates for the current set of actions can be effectively used to estimate values of a new set of actions.

%Aditi : Changed conclusion
\section{Conclusion}
We have presented RLDT, an RL-based online algorithm that actively chooses to know very few features of a data point to \textit{both} classify it and to learn a better model.
The framework of the algorithm is rich enough to handle different settings such as class imbalance and non-uniformly expensive features using appropriate parameter values. Furthermore, since we can employ any RL algorithm to learn the optimal policy, RLDT promises multiple solution methods to online decision tree learning that can be explored in the future.

\bibliography{RLDT}

\end{document}